# Identification of Books That are Suitable for Middle School Students Using Artificial Neural Networks

Sadık Ozan Görgü[1], Ege Gençer[1], Alp Niksarlı[2],

[1]Robert College, İstanbul, Turkey

[2]Department of Computer Engineering, Faculty of Engineering, Davidson College, North Carolina, America

**Abstract:** Reading right books contributes to children's imagination and brain development, enhances their language and emotional comprehension abilities, and strengthens their relationships with others. Building upon the critical role of reading books in individual development, this paper aims to develop an algorithm that determines the suitability of books for middle school students by analyzing their structural and semantic features. Using methods described, an algorithm will be created that can be utilized by institutions and individuals responsible for children's education, such as the Ministry of National Education officials and schools. This algorithm will facilitate the selection of books to be taught at the middle school level. With the algorithm, the book selection process for the middle school curriculum can be expedited, and it will serve as a preliminary reference source for those who evaluate books by reading them. In this paper, the Python programming language was employed, utilizing natural language processing methods. Additionally, an artificial neural network (ANN) was trained using the data which had been preprocessed to construct an original dataset. To train this network, suitable books for middle school students were provided by the MEB, Oxford and Cambridge and with content assessed based on the "R" criterion, and inappropriate books for middle school students in terms of content were included. This trained neural network achieved a 90.06% consistency rate in determining the appropriateness of the test-provided books. Considering the obtained findings, it can be concluded that the developed software has achieved the desired objective.

## 1. Introduction

With the expansion of the Web and its easy accessibility, in the 21st century, the transfer of information has rapidly and sometimes uncontrollably increased. This expansion has had its effects on education as well. Although at first these new technologies (e.g., Smartboards, computers) were approached with caution, educators in the world have come to realize its complimentary role (Raja & Nagasubramani, 2018). Since the sudden expansion of chatbots (e.g., ChatGPT), some educators have been complaining about the ease technology brings the students and think it will be harmful for students; on the other hand, some think it is a source that can be very helpful. Although there are restrictions on the content the chatbots can generate, it hasn't always this way. Also, on the Web students face the possibility of exposure to inappropriate content. One of the most important contents middle school children face is their school books and the complimentary reading books. The objective of this paper is to examine sentences within children books from both structural and semantic perspectives using natural language processing (NLP) techniques in order to identify those that could contribute to the personal development of children because exposure of children to inappropriate content for their age has numerous adverse effects. Consequently, through this study, it will be possible to mitigate the potential negative impacts of books containing inappropriate content on children. Additionally, this algorithm aims to mitigate personal and societal biases that significantly influence evaluations made by individuals, by generating numerical results.

The main contribution of this study is as follows:

• To the best of our knowledge, it is the first study to analyze books themselves with sentiment analysis to find their suitability.

The remainder of this paper is organized as follows. In Section 2, we briefly discuss related work on sentiment analysis. Section 3 presents the preprocessing stage and unique dataset created after the preprocessing stage. In Section 4 we consider the experimental study, such as creating the ANN based on the unique dataset. In Section 5 we evaluate our ANN model by comparing the results given by the trained model for unseen data and the real suitability values. Finally, in Section 6, we discuss the conclusion and future work.

## 2. Related work

While the use of sentiment analysis techniques has become more widespread, these techniques are primarily employed for analyzing social media comments or comments related to products. However, the paper written by Shaik & et.al (2023) encompasses four distinct levels of emotional analysis in education: document level, sentence level, entity level, and aspect level. Additionally, it delves into both lexicon-based and corpus-based approaches for unsupervised annotations. The article further elucidates the pivotal role of AI in sentiment analysis, highlighting methodologies such as machine learning, deep learning, and transformers.

On the other hand, in studies that based on social media comments or comments related to products, comments are generally subjected to sentiment analysis in three different ways: at the text level, sentence level, and word level (Liu, 2012). When examining social media comments, analysis is often conducted at the word level. For instance, in their analysis of Twitter comments, Gann et al. (2014) assigned a value called Total Sentiment Value (TSV) to each symbol (word) and thus categorized words as positive or negative. TSV is calculated as shown below:

$$\text{TDD} = \frac{p - \frac{tp}{tn} \times n}{p + \frac{tp}{tn} \times n}$$

where p represents the number of positive sentences the word appears in, n represents the number of negative sentences the word appears in, tp represents the total number of positive sentences in the text, and tn represents the total number of negative sentences in the text. When examining social media comments, either machine learning methods, dictionary-based analysis methods, or a combination of the two are commonly used. Both of these methods have advantages and disadvantages that have been revealed in previous studies. For example, machine learning methods like SVM and Naïve Bayes models, while capable of understanding the different meanings of words based on their usage in text, often require more time (Drus and Khalid, 2019). On the other hand, the Naïve Bayes model, although capable of being trained with less data, yields inconsistent results in the presence of spelling errors in Facebook comments (Akter and Aziz, 2016). Similar techniques are employed when analyzing product or service reviews. Martineau and Finin (2009) classified reviews as positive or negative using the Delta TF-IDF function to examine comments.

Product reviews generally include a rating ranging from 1 to 5 or 1 to 10. These ratings are taken as reference values, allowing algorithms to be trained and their consistency tested. De Albornoz et al.'s (2011) study assigns scores to reviews by identifying words such as "room," "location," and "staff" in hotel reviews. This approach improves the understanding of reviews and results in more consistent and objective ratings (De Albornoz et al., 2011).

Emotional analysis techniques are widely used in the business sector and social sciences due to their importance in terms of time savings and provided data (O'Mahony et al., 2010). However, this method is not extensively used in the education sector. One of the few studies conducted in the field of education focuses on children's books. The closest study related to children's books involves performing emotional analysis of book reviews to determine their suitability for children (Choi, 2019).

Upon reviewing the existing literature, it can be observed that these studies examine books based on evaluations made by readers online. The main challenge with this method is the discrepancy between the given scores and the written comments. For instance, a review receiving a score of 7 out of 10 could contain a comment such as "The hotel looks quite outdated. The breakfast room isn't large enough to accommodate big breakfasts" (Maks and Vossen, 2013). A potential customer reading this comment might choose not to opt for the mentioned hotel, despite the 7 out of 10 score. This inconsistency arises from the fact that the individual providing the online review shares their experience and scores their experience, resulting in such discrepancies (Maks and Vossen, 2013). In summary, a significant majority of the studies in this field so far make inferences about products by analyzing the experiences and opinions of users. Analyzing the products themselves, focusing on the concrete features of the product rather than abstract opinions of users, could lead to more consistent and accurate results. In this regard, we believe that our study will contribute to positive developments, particularly in the education sector and various other fields.

## 3. Materials and methods

Figure 1 shows the proposed method that aims to generate a neurol network which is be able to classify middle school children's books depending on their suitability. We apply six main steps, which consist of data collection, preprocessing, feature extraction, feature generation, constructing ann, and obtaining the results.

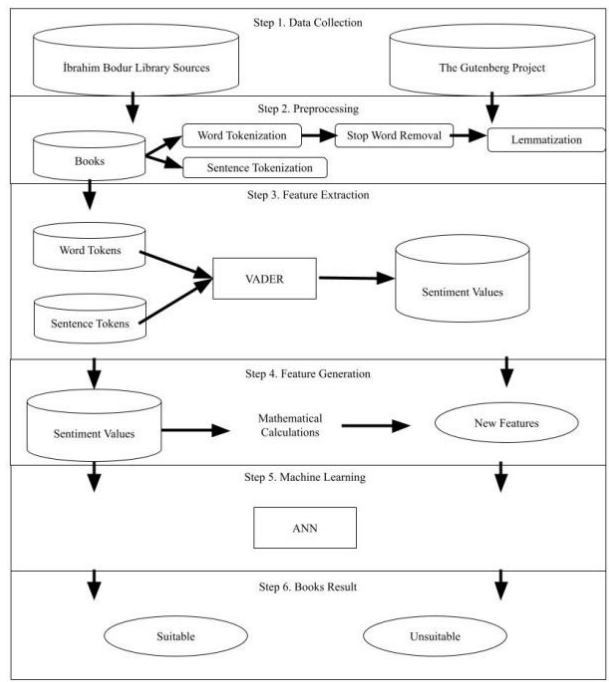

**Figure 1**. Our proposed framework

### 3.1. Unique Dataset

Due to the lack of a database on which the project can be based, a unique dataset was created. This dataset contains both suitable and unsuitable books for middle school children. We used children books approved by the Turkish Ministry of National Education and other sites such as Oxford and Cambridge University Libraries as suitable books. On the other hand, for the unsuitable books, we used books that was given R-rating, in which sexual, horror, alcohol, and drug content was existing. Furthermore, we used Amazon's age ratings to determine the suitability of several books. At the end, our dataset stores the information of 416 books.

Table 1. Features and their selection reasons

| Features | Reasons |
| --- | --- |
| Number of positive sentences | For the books with similar numbers of total sentences, the number of positive and negative sentences can be used to guess the suitability of the book, since suitable books have more positive sentences. |
| Number of negative sentences | |
| Number of total sentences | For the books with different total sentences, ratios become more critical. In order to find the ratios number of total sentences must be recorded. |
| Ratio of positive sentences to total sentences | For the books with different total sentences, ratios play a more important role in determining the suitability of the books. |
| Ratio of positive sentences to negative sentences | |
| Number of positive words | The same reasons indicated above for the sentences but in this case for words |
| Number of negative words | |
| Number of total words | |
| Ratio of positive words to total words | |
| Ratio of positive words to negative words | |
| Words per sentence | Elementary school books tend to have fewer words per sentence whereas more complicated novels tend to have higher words per sentence |
| Ratio of middle-school words to total words | These parameters are used to understand the complexity of the text |
| Ratio of high-school words to total word | |

| Coefficient of positivity | Refer to section 3.3 |
|---|---|

It's relatively a small number for a dataset; it's because of the copyrights. Due to copyright of the books, we were only able to use our school's library sources and the books in "The Gutenberg Project", which is a website that includes non-copyrighted books. In this dataset, there are 16 information are stored. Name of the book, 14 parameters and suitability tag (0: unsuitable, 1: suitable).

Table 2. The unique dataset.

| Book Name | # of Positive Sentences | # of Negative Sentences | # of Total Sentences | Positive Sent. : Total Sent. | Positive Sent. : Negative Sent. | # of Positive Words | # of Negative Words | # of Total Words | Positive Words : Total Words | Positive Words : Negative Words | Words per Sentence | %Middle School Words | %High School Words | Coefficient of Positivity | Suitability |
|---|---|---|---|---|---|---|---|---|---|---|---|---|---|---|---|
| Gullivers Travels | 1303 | 697 | 2506 | 0.5199521149 | 1.869440459 | 4391 | 2501 | 104972 | 0.04183020234 | 1.755697721 | 41.88826816 | 6.09E-05 | 0.01270498458 | 52.83109948 | 1 |
| 1984 | 1662 | 1827 | 5896 | 0.2818860244 | 0.9096880131 | 3679 | 3253 | 102758 | 0.03580256525 | 1.130956041 | 17.42842605 | 0.0001216224434 | 0.007317617011 | 10.75150142 | 0 |
| Rose in Bloom, by Louisa May Alcott | 1974 | 746 | 3268 | 0.6040391677 | 2.646112601 | 6997 | 2732 | 103088 | 0.06787404936 | 2.561127379 | 31.54467564 | 0.0001779113211 | 0.006899005673 | 76.34080467 | 1 |
| Kate's Ordeal, by Emma Leslie | 361 | 268 | 1080 | 0.3342592593 | 1.347014925 | 943 | 548 | 21342 | 0.04418517477 | 1.72080292 | 19.76111111 | 0 | 0.002581926514 | 70.28238875 | 1 |
| Fevre Dream | 2790 | 3063 | 11642 | 0.2396495448 | 0.9108716944 | 5185 | 4735 | 139465 | 0.03717778654 | 1.095036959 | 11.97947088 | 0.0005009159606 | 0.003721089993 | 4.686050053 | 0 |
| Robin Hood | 2308 | 1146 | 4993 | 0.462247146 | 2.013961606 | 5432 | 2565 | 112425 | 0.04831665555 | 2.117738791 | 22.51652313 | 7.67E-04 | 0.002778819598 | 42.24364462 | 1 |
| Happy-prince-and-other-tales | 344 | 211 | 768 | 0.4479166667 | 1.630331754 | 867 | 428 | 16421 | 0.05279824615 | 2.025700935 | 21.38151042 | 0 | 0.0009350788138 | 58.63310123 | 1 |
| Fifty Shades Darker | 4723 | 3335 | 15056 | 0.3136955367 | 1.416191904 | 9366 | 5539 | 163868 | 0.05715575951 | 1.690918938 | 10.88390011 | 2.63E-04 | 0.006775032168 | 10.53868387 | 0 |
| The Loop | 1486 | 1958 | 6005 | 0.2474604496 | 0.7589376915 | 3587 | 3509 | 91496 | 0.03920389962 | 1.022228555 | 15.23663614 | 0.0002274847017 | 0.00416297004 | 3.812179899 | 0 |
| Heidi | 1332 | 669 | 3186 | 0.418079096 | 1.99103139 | 2939 | 1348 | 66818 | 0.0439851537 | 2.180267062 | 20.97237916 | 3.28E-05 | 0.001182499015 | 67.82475257 | 1 |
| The Attack in Trench Warfare: of a Company Commander | 182 | 333 | 677 | 0.2688330871 | 0.5465465465 | 613 | 916 | 18522 | 0.03309577799 | 0.6692139738 | 27.35893648 | 0 | 0.009215646119 | 7.109295007 | 0 |
| Happy Jack, by Thornton Burgess | 776 | 320 | 1621 | 0.4787168415 | 2.425 | 1511 | 676 | 26502 | 0.05701456494 | 2.235207101 | 16.34916718 | 0 | 0.0007795447459 | 75.3509872 | 1 |
| Pollyanna | 1369 | 878 | 3582 | 0.3821887214 | 1.559225513 | 3015 | 1723 | 60590 | 0.04976068658 | 1.749854904 | 16.91513121 | 3.48E-05 | 0.002191685511 | 38.6624638 | 1 |
| The Wonderful Wizard of Oz | 912 | 620 | 2400 | 0.38 | 1.470967742 | 1819 | 1411 | 41365 | 0.04397437447 | 1.289156627 | 17.23541667 | 0.0001018381791 | 0.0004582718061 | 44.85928952 | 1 |
| bird_box | 1425 | 1495 | 8029 | 0.1774816291 | 0.9531772575 | 2414 | 1917 | 71740 | 0.0336492891 | 1.259259259 | 8.935110225 | 7.72E-04 | 0.002115857494 | 7.285318077 | 0 |
| the-devil-all-the-time | 1107 | 1660 | 4875 | 0.2270769231 | 0.6668674699 | 2523 | 2794 | 88148 | 0.02862231701 | 0.9030064424 | 18.08164103 | 7.05E-05 | 0.001174315374 | 1.892830047 | 0 |
| my-sweet-orange-tree | 1039 | 779 | 3382 | 0.3072146659 | 1.333761232 | 1905 | 1249 | 45254 | 0.04209572634 | 1.525220176 | 13.38083974 | 1.00E-04 | 0.001199820027 | 28.32203676 | 1 |
| Shutter Island | 1271 | 1406 | 6014 | 0.2113402062 | 0.9039829303 | 2693 | 2328 | 85013 | 0.03167750815 | 1.156786942 | 14.13584968 | 0.0002452844073 | 0.003581152346 | 2.459686392 | 0 |
| Chronicles_of_narnia | 6181 | 5175 | 23285 | 0.2654498604 | 1.194396135 | 12073 | 8111 | 328982 | 0.03669805643 | 1.488472445 | 14.12849474 | 0.0001197788085 | 0.001483926349 | 23.75889448 | 1 |

In these 14 parameters, as it can be seen from Table 1 there are both sematic and structural information regarding books. First of all, number of positives sentences, number of negative sentences, and number of total sentences are recorded. The reason why they are recorded is that for the books with similar number of total sentences, the higher the number of positive sentences, higher the chance that it can be a suitable book. This relationship can be also hypothesized between number of negative sentences and suitability: The higher the number of negative sentences, higher the chance that it can be a unsuitable book. However, for books with different lengths, these numbers can be deceiving. For instance, a short suitable story might have fewer positive sentences than an extremely long unsuitable book. To avoid this problem, there are also ratio-based parameters: ratio of positives sentences to total sentences and ratio of positive sentences to negative sentences. These 5 parameters for sentences are also applied for words (number of positive words, number of negative words, number of total words, ratio of positive words to total words, and ratio of positive words to negative words) due to the same reasons. Then, there is "Words per Sentence" parameter. The reason for this parameter is that most of the elementary-level books have shorter sentences, whereas more technical or high-level texts have greater words per sentence. We also created "middle-school words" list and "high-school words" list by using the datasets of Membean. Membean is a vocabulary program, in which there are approximately 4500 words that are mostly used per each level. By using these word lists, we also calculated the ratio of middle-school words to total words and the high-school words to total words and stored them as parameters. The main reason why we used them as parameter is to detect the complexity of the text. Lastly, the 14th parameter is "coefficient of positivity, which will be explained in 3.3. As it can be seen from

Table 2, previously mentioned 14 parameters are used to create a unique dataset which in turn is used to train the neural network.

### 3.2. Preprocessing

While developing our algorithm, we utilized various libraries: Numpy, NLTK Toolkit, Keras, and Scikit-learn. For sentiment analysis to be conducted, textual data must undergo a pre-processing stage. This pre-processing stage involves tokenization, vectorization, lemmatization, and the removal of unnecessary characters (stopword removal).

Initially, to simplify the text and eliminate redundant spaces between letters and words that arise while converting text from PDF to plain text format, these spaces are replaced with regular (single) spaces.

In this study, tokenization is performed both at the word and sentence levels, meaning that the text is segmented based on both sentences and words. For the tokenization at the word level Regexp Tokenizer is used. During tokenization at the word level, words that do not influence the emotional value of a sentence, such as pronouns, conjunctions, and prepositions, are filtered out.

Following word filtering, lemmatization is employed to simplify words by removing their suffixes and transforming them to their base forms. For lemmatization WordNet Lemmatizer is used. Through lemmatization, words are stemmed based on their contextual usage within the sentence.

On the hand, tokenization at the sentence level takes into consideration the meanings derived from the context of words within sentences. This approach enhances the precision of sentiment analysis. For this tokenization process Punkt Sentence Tokenizer is used.

### 3.3. Feature extraction with sentiment analysis (VADER)

After passing through the text pre-processing stage, the sentiment values of the sentences and words comprising the text are determined by using VADER, which is developed by Hutto & Gilbert (2014). VADER is a widely employed tool for sentiment analysis in natural language processing. It proves particularly useful for analyzing social media texts and other informal languages due to its ability to understand and interpret slang, emotions, and complex sentence structures. An advantage of VADER over other sentiment analysis tools is its capability to accurately detect irony and sarcastic expressions within the text.

Since VADER is mainly used to determine the positivity and negativity of the text, sometimes it is not able to detect inappropriate contents for the middle school children. To avoid this problem, we created a "Bad-Words" list, in which approximately 1700 words that are considered unsuitable for children due to reasons such as racial slurs, sexual content, profanity, and slang are stored. If a word from the "Bad-Words" list is present in the text, the sentiment value of that word is updated to -1.0, regardless of the value assigned by VADER.

A similar process is carried out at the sentence level. Sentences containing words that are deemed inappropriate for children are automatically assigned a sentiment score of -1.0 by VADER. This process ensures that the sentiment scores assigned by VADER to sentences with inappropriate words are not influenced by other words in the sentence. In other words, this step prevents the sentiment impact of an inappropriate word within a sentence from being reduced due to the influence of other words. Essentially, this process is implemented because we consider that just one "highly inappropriate" word is sufficient to render an entire sentence inappropriate.

Following the categorization of symbols by VADER into positive, neutral, and negative sentiments, an elimination process is applied to focus on positive and negative symbols rather than neutral ones. This is due to the fact that the primary determinants of a text's sentiment are the positive and negative symbols. To achieve this, the sentiment values assigned to symbols by VADER, representing the symbol's emotional intensity, are employed. Based on these values, sentences are divided into two separate lists: positive and negative. These positive and negative sentences from the books are utilized to calculate coefficient of positivity which is one of the 14 parameters then used to train the model.

In his work titled "Finding Just Right Books for Children: Analyzing Sentiment in Online Book Reviews," Choi (2019) proposes that if the absolute sentiment value of a symbol is greater than 0.8, then the consistency of that symbol's sentiment is high, suggesting that greater importance should be given to symbols with such values. On the other hand, Choi (2019) also asserts in the same study that symbols with an absolute sentiment value below 0.5 exhibit weaker consistency and hence have a lesser impact on the text. This assertion is because even though a book is generally positive, a few sentences containing inappropriate content such as violence, drugs, and sexual themes can render the entire book inappropriate. To address this, 14th power of the sentiment values are taken. This process allows sentences with absolute values greater than 0.8 to be weighted more significantly in the calculated total sentiment value by being multiplied by a higher value.

**Figure 2**. Coefficient of positivity graph.

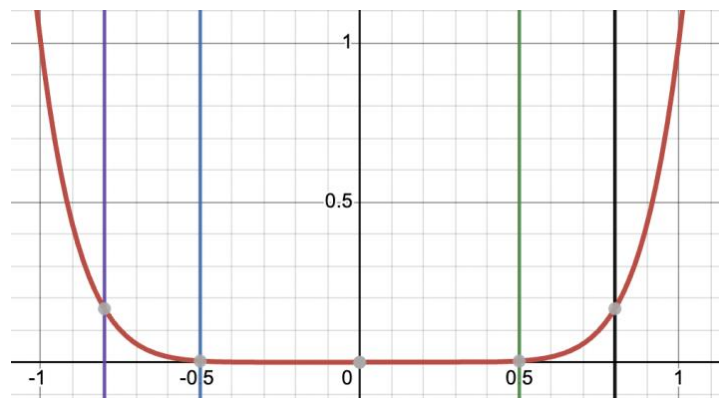

Consequently, sentences with absolute values exceeding 0.8 have a greater influence on determining a book's appropriateness compared to sentences with values between -0.5 and 0.5. The combination of VADER's semantic evaluation, sentences containing violence, alcohol, and sexual content receiving sentiment values lower than -0.8, and the explicit assignment of sentiment values of -1.0 to profane, slang, and sexually explicit words) enhances the effectiveness of this code, making sentences with absolute values exceeding 0.8 a key factor in determining a book's appropriateness. As depicted in Figure 2, $14^{th}$ power ensures that the area under the curve remains entirely positive. Thus, the calculated coefficient of positivity will always be positive, regardless of how large the sentiment value from negative sentences is.

At this point, the percentage of the calculated coefficient of positivity, which determines the positivity contributed by sentiment values of positive sentences to assess the book's appropriateness, is calculated. This calculated coefficient of positivity serves as one of the 14 parameters used to compute the book's appropriateness.

### 3.4. Artificial neural networks

Artificial Neural Networks (ANNs), a machine learning model inspired by the structure and functionality of biological neural networks, are composed of layers of "neurons" that process and transmit information (Grossi & Buscema, 2007). Among the most commonly used types of ANNs in natural language processing (NLP) is the Recurrent Neural Network (RNN), a type of ANN particularly useful for tasks involving sequential data, such as language modeling and machine translation. For instance, in the paper titled "Effective Approaches to Attention-Based Neural Machine Translation" (Bahdanau et al., 2015), the authors propose an attention-based artificial neural network machine translation system that employs one RNN for encoding the source sentence and another RNN for decoding the target sentence. This approach demonstrates superior performance compared to traditional pattern-based machine translation systems across various language pairs.

In general, ANNs have proven highly effective in tasks related to text generation, sentiment analysis, language modeling, and a wide range of other NLP tasks, making them widely adopted in numerous applications.

## 4. Experimental studies
### 4.1. Training the neurol network on the unique dataset

The neural network we created consists of 4 different layers: input, output and 2 hidden. While creating the neural network, different hyperparameters were validated and optimal ones were selected. The input size was chosen as 14 since there are 14 input variables. Used activation functions were relu and sigmoid for hidden layers and adam optimization function for the output layer.

**Figure 3**. Hyperparameter tuning.

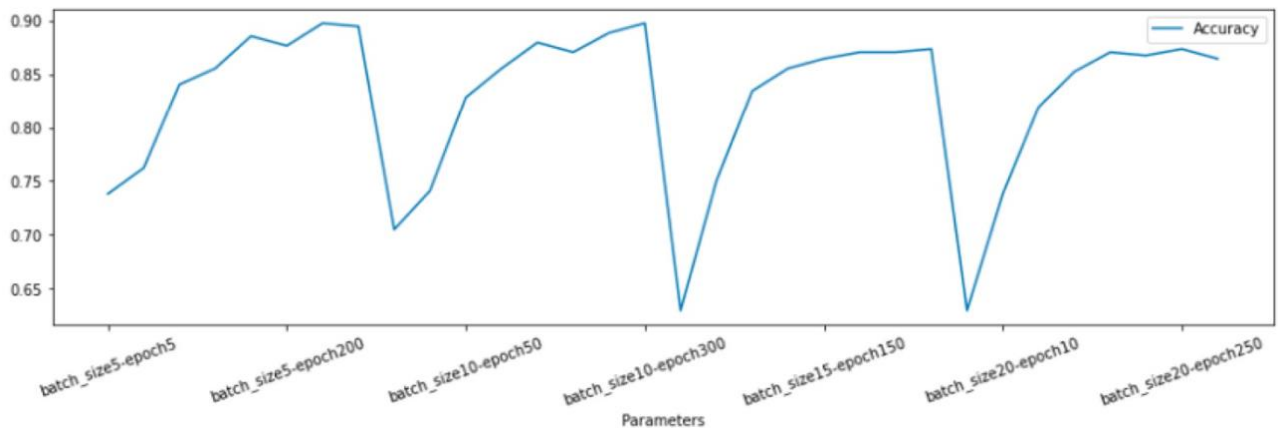

Hyperparameter tuning is of great importance in order to increase the accuracy of the artificial neural network. In order to select the optimal values, the function seen in Figure 3 is used. This function tests different hyperparameters using multiple loops and saves the accuracy values for each. The recorded values are reflected in Figure X. Based on the accuracy values of the batch size and epochs numbers observed here, the actual artificial neural network values have been updated and optimized.

### 4.2. The confusion matrix

Table 3. Definition of confusion matrix.

| | | Predicted class | |
|---|---|---|---|
| | | P | N |
| Actual class | P | TP (True positives): The number of true positives, i.e. the number of files that are classified as positive correctly | FN (False negatives): The number of false negatives, i.e. the number of files that are classified as negative incorrectly |
| | N | FP (False positives): The number of false positives, i.e. the number of files that are classified as positive incorrectly | TN (True negatives): The number of true negatives, i.e. the number of files that are classified as negative correctly |

Accuracy (Acc) is the ratio of the number of documents that are correctly classified to the total number of documents. The calculation of accuracy is given in Eq. (1).

$$Acc = (TP + TN)/(TP + TN + FP + FN) \quad (1)$$

Precision (Pr) is the probability that a randomly selected document is retrieved as relevant. It is calculated as the ratio of the total number of positive files that are correctly classified to the total number of positive classified files, as in Eq. (2):

$$Pr = TP/(TP + FP) \quad (2)$$

Recall (Re) is the probability that a randomly selected relevant document is retrieved in a search. It is calculated as the ratio of total number of positive files that are correctly classified to the number of positive files that are in the dataset, as in Eq. (3):

$$Re = TP/(TP + FN) \quad (3)$$

The F-measure (Fm) is the harmonic mean of precision and recall and it is calculated as in Eq. (4):

$$Fm = 2*Pr*Re/(Pr+Re) \quad (4)$$

## 5. Experimental results

When examining the findings, it is evident that the coefficient of positivity value of a book serves as an informative measure regarding the suitability of the book for middle school children. Taking the value of 30 for the coefficient of positivity, out of the 416 books present in the dataset, 311 are accurately classified. In other words, with a coefficient of positivity value set at 30, it achieves an accuracy of 74.7%.

Table 4. The confusion matrix for the predicted books

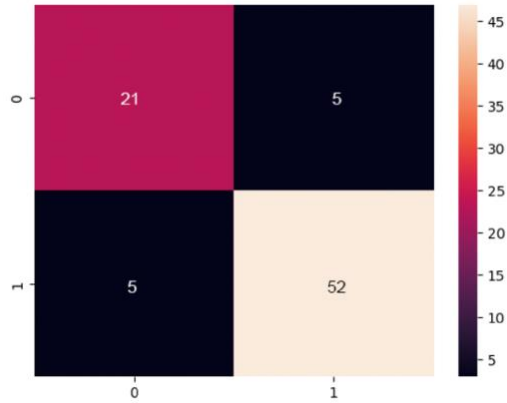

The artificial neural network (ANN) model developed with 416 books demonstrates the ability to successfully detect appropriate books with an accuracy of 91.2% and identify inappropriate books with an accuracy of 80.7% when tested against a sample dataset of 83 books, as it can be seen from Table 4. Overall, the model exhibits a consistency rate of 90.06%, attributed to its training with a relatively large dataset. Table 5 provides examples of well-known books which the model has been tested on.

Table 5. Output of the ANN model and comparison.

| Name of the Book | Predicted Suitability | Actual Suitability |
|---|---|---|
| My Sweet Orange Tree | 0.967 | 1 |
| 1984 | 0.022 | 0 |
| Women | 0.021 | 0 |
| Calliou | 0.786 | 1 |
| Awakening | 0.251 | 0 |
| Alice in Wonderland | 0.861 | 1 |
| Wealth of Nations | 0.070 | 0 |
| Harry Potter | 0.833 | 1 |
| Around the World in 80 Days | 0.930 | 1 |
| Metamorphosis | 0.159 | 0 |
| The Drawing of the Three | 0.001 | 0 |

## 6. Conclusion and future work

When taking the findings above into account, it can be seen that coefficient of positivity can be utilized as an input in the artificial neural network model. Although it's not enough to use this value solely to determine the suitability of the books, it can be useful to use as a parameter.

The results reveal that among books with coefficient of positivity values higher than 30%, which are deemed inappropriate for middle school children, complexly worded texts laden with jargon (such as academic articles) are prominent. Additionally, it can be inferred that creating an academic jargon list and incorporating it into the code is necessary for a more effective operation of the model. Furthermore, examining the coefficient of positivity values of books categorized as unsuitable for middle school children due to elements of violence, sexuality, fear, and substance usage indicates that the coefficient of positivity value of 30 consistently functions accurately, demonstrating the correct identification of these elements within the books.

To enhance the accuracy and consistency of the developed model, it is imperative to expand the database used in its training. In other words, testing this ANN model with at least 3000 books, rather than 416, could yield more consistent outcomes.

The developed ANN model in this study encounters difficulties in distinguishing academically complex articles that are comprehensible to middle school children. Although this ANN model is successful in promptly identifying the presence of violence, cruelty, sexuality, addiction, and fear elements, it struggles to recognize complex jargon in academic works. We have included over 3000 words most commonly used in daily conversations to be able to understand the complexity of a text. Another solution is that by tailoring the model to the domains of academic articles, extracting frequently used terms, and creating a new dataset for training, can improve the model.

With the completion of the model's deficiencies, the developed website can be employed by the Ministry of Education, middle schools, teachers, parents, and students to assess the suitability of books for middle school-level children.

"The Oxford Children's Corpus" (OCC) project has been requested from Oxford University, but as of now, there has been no response from them. The utilization of this corpus in obtaining and training the ANN module could potentially increase the achieved consistency rate of 83.3%. Similarly, by expanding the number of books deemed inappropriate for middle school children, the ANN module could be trained to yield more realistic results.

Our software, with the establishment of an internet website within the Ministry of Education (MEB), could be employed as a screening tool for selecting books to be taught in schools. Priority can be given to books with positivity values below 10%, and books with positivity values exceeding 87% can be directly incorporated into the curriculum. This approach can expedite the book selection process and provide readers with a new evaluation criterion they can use as a reference.

By increasing the number of books categorized according to age ranges in the dataset, our software can output the age group to which the entered book belongs. The forthcoming internet website can present users, whether students or teachers, with information about which age groups the books, either entered as ".txt/.pdf" files or stored in the database, are suitable for.